\documentclass[11pt,a4paper]{article}
\usepackage[hyperref]{eacl2021}
\usepackage{times}
\usepackage{latexsym}
\usepackage{hyperref}
\makeatletter
\newcommand{\printfnsymbol}[1]{%
  \textsuperscript{\@fnsymbol{#1}}%
}
\makeatother

\usepackage{booktabs}
\usepackage{amsmath}
\usepackage{graphics}
\usepackage{graphicx}
\usepackage[normalem]{ulem}

\usepackage{multirow}
\usepackage[normalem]{ulem}
\useunder{\uline}{\ul}{}

\usepackage{microtype}

\aclfinalcopy 

\newcommand{\namedref}[2]{\hyperref[#2]{#1~\ref*{#2}}}
\newcommand{\sectionref}[1]{\namedref{Section}{#1}}
\newcommand{\tableref}[1]{\namedref{Table}{#1}}
\newcommand{\figureref}[1]{\namedref{Figure}{#1}}
\newcommand{\appendixref}[1]{\namedref{Appendix}{#1}}

\title{Long Document Summarization in a Low Resource Setting \\ using Pretrained Language Models}

\author{Ahsaas Bajaj\thanks{* Equal Contribution} \\
  University of Massachusetts Amherst \\
  \texttt{abajaj@umass.edu} \\
  
 \And
 
 Pavitra Dangati\printfnsymbol{1}  \\
  University of Massachusetts Amherst \\
  \texttt{sdangati@umass.edu} \\
   
  \AND
 
 Kalpesh Krishna \\
  University of Massachusetts Amherst \\
  
  \And
  
   Pradhiksha Ashok Kumar \\
  University of Massachusetts Amherst \\
  
  \AND
  
  Rheeya Uppaal \\
  Goldman Sachs \\
  
  \And
  Bradford Windsor \\
  Goldman Sachs \\
  \And
  Eliot Brenner \\
  Goldman Sachs \\
  \And
  
  Dominic Dotterrer\\
  Goldman Sachs \\
  \AND
  Rajarshi Das \\
  University of Massachusetts Amherst \\
  \And
  Andrew McCallum \\
  University of Massachusetts Amherst \\
  }

\date{}

\begin{document}
\maketitle
\begin{abstract}
 
Abstractive summarization is the task of compressing a long document into a coherent short document while retaining salient information. Modern abstractive summarization methods are based on deep neural networks which often require large training datasets. Since collecting summarization datasets is an expensive and time-consuming task, practical industrial settings are usually \emph{low-resource}. In this paper, we study a challenging low-resource setting of summarizing long legal briefs with an average source document length of 4268 words and \textbf{only 120} available (document, summary) pairs. To account for data scarcity, we used a modern pretrained abstractive summarizer BART~\citep{lewis2019bart}, which only achieves 17.9 ROUGE-L as it struggles with long documents. We thus attempt to compress these long documents by identifying salient sentences in the source which best ground the summary, using a novel algorithm based on GPT-2~\citep{radford2019language} language model perplexity scores, that operates within the low resource regime. On feeding the compressed documents to BART, we observe a 6.0 ROUGE-L improvement. Our method also beats several competitive salience detection baselines. Furthermore, the identified salient sentences tend to agree with an independent human labeling by domain experts.


\end{abstract}

\begin{figure*}[ht!]
    \centering
    \includegraphics[scale=0.53]{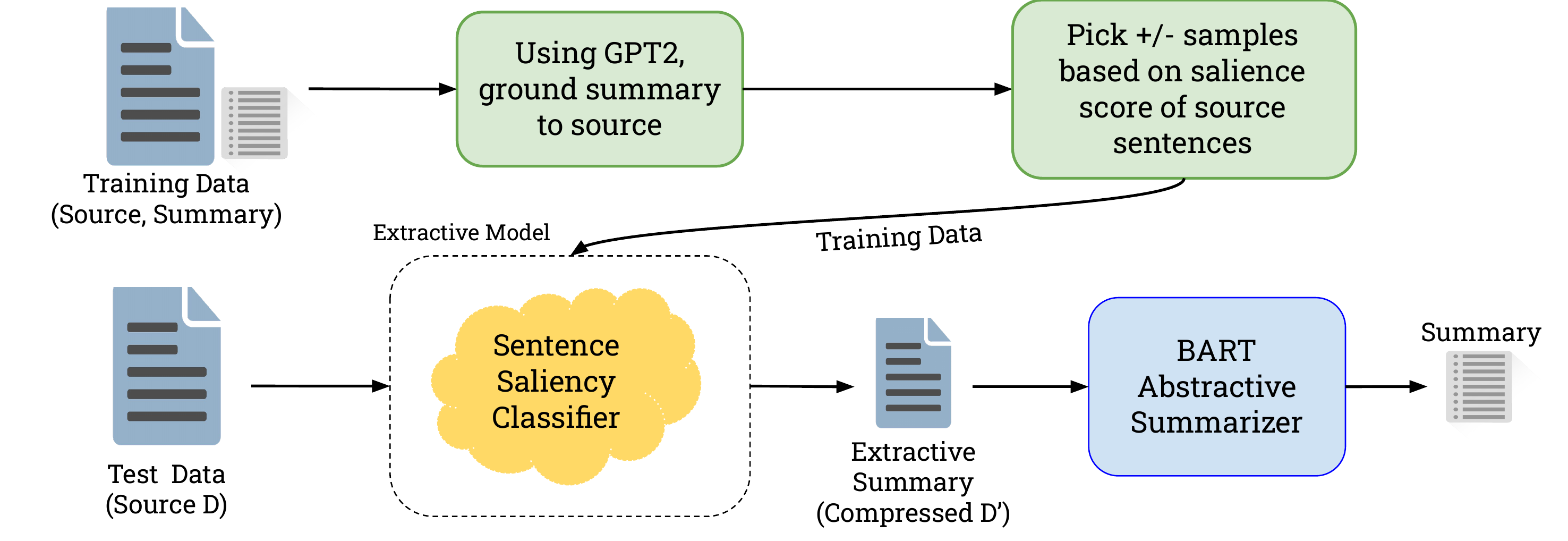}
    \caption{Our method for long document summarization task in low resource setting. The \textit{Extraction Model} generates a compressed document $D'$ by identifying salient sentences. It is trained by computing salience score for each training set source sentence. The pretrained abstractive summarizer takes as input the compressed document.}
    \vspace{-0.1in}
    \label{fig:my_label}
\end{figure*}
\section{Introduction and Related Work}




Text summarization is the task of generating a smaller coherent version of a document preserving key information. Typical abstractive summarization algorithms use seq2seq models with attention \citep{chopra-etal-2016-abstractive}, copy mechanisms \citep{gu-etal-2016-incorporating}, content selection \citep{cheng-lapata-2016-neural}, pointer-generator methods \citep{See_2017} and reinforcement learning \citep{wu2018learning}. These methods perform well in high resource summarization datasets with small documents such as CNN/DailyMail~\citep{nallapati-etal-2016-abstractive}, Gigaword~\citep{Rush_2015}, etc. However, summarization over long documents with thousands of tokens is a more practically relevant problem. Existing solutions focus on leveraging document structure \cite{cohan-etal-2018-discourse} or do mixed model summarization involving compression or selection followed by abstractive summarization~\citep{liu2018generating, gehrmann-etal-2018-bottom}. However, these methods require large amounts of training data. Low resource settings are common in real world applications as curating domain specific datasets especially over long documents and on a large scale, is both expensive and time consuming.

A human summarizing a long document would first understand the text, then highlight the important information, and finally paraphrase it to generate a summary. Building on this intuition, we present a low-resource long document summarization algorithm (\sectionref{sec:approach}) operating in 3 steps --- \\\textbf{(1)} ground sentences of every training set summary into its source, identifying salient sentences;\\\textbf{(2)} train a salience classifier on this data, and use it to compress the source document during test time;\\ \textbf{(3)} feed the compressed document to a state-of-the-art abstractive summarizer pretrained on a related domain to generate a coherent and fluent summary.

To tackle data scarcity, we use pretrained language models in all three steps, which show strong generalization~\citep{devlin-etal-2019-bert} and are sample efficient~\citep{yogatama2019learning}. Notably, our step \textbf{(1)} uses a novel method based on GPT-2 perplexity~\citep{radford2019language} to ground sentences. 

Unlike prior work~\citep{parida-motlicek-2019-abstract, magooda2020abstractive} tackling data scarcity in summarization, our method needs no synthetic data augmentation. Moreover, we study a significantly more resource constrained setting --- a complex legal briefs dataset (\sectionref{sec:approach}) with \textbf{only 120} available (document, summary) pairs and an average of 4.3K tokens per document;~\citet{parida-motlicek-2019-abstract} assume access to 90,000 pairs with a maximum of 0.4K source document tokens,~\citet{magooda2020abstractive} use 370 pairs with 0.2K source document tokens.

Despite this challenging setup, our method beats an abstractor-only approach by 6 ROUGE-L points, and also beats several competitive salience detection baselines (\sectionref{sec:experiments}). Interestingly, identified salient sentences show agreement with an independent human labeling by domain experts, further validating the efficacy of our approach.




\label{sec:dataset}

 
\section{Dataset and Approach}
\label{sec:approach}

To mimic the real world scenario of summarization over long domain-specific documents, we curate 120 document-summary pairs from publicly available Amicus Briefs\footnote{\url{https://publichealthlawcenter.org/amicus-briefs}}, thus simulating the legal domain.\footnote{The source contains detailed arguments that the court should consider for a case; the target summarizes them.} As shown in \tableref{tab:dataset}, our dataset is significantly smaller than the popular CNN/Daily Mail benchmark~\citep{nallapati-etal-2016-abstractive} and has significantly longer documents and summaries.

\begin{table}[h]
\centering
\small
\begin{tabular}{lrrr}
\toprule
Dataset & \# $(S, T)$ & Avg. $|S|$ & Avg. $|T|$ \\
\midrule
CNN/DM & 312084 & 781 & 56\\
Amicus & 120 & 4268 & 485 \\
\bottomrule
\end{tabular}
\vspace{-0.05in}
\caption{A comparison between the Amicus legal briefs dataset and the popular CNN/Daily Mail benchmark. Amicus has far fewer document-summary pairs \#$(S, T)$, with more documents tokens (Avg. $|S|$) and summary tokens (Avg. $|T|$) on average.}
\label{tab:dataset}
\vspace{-0.1in}
\end{table}
 
To tackle this low resource setting, we use the state-of-the-art abstractive summarizer BART~\citep{lewis2019bart}, pretrained on a dataset from a related domain (CNN/DM). Since BART was trained on short documents, it truncates documents longer than 1024 subwords. Hence, instead of feeding the whole source document as input to BART, we feed salient sentences extracted using a salience classifier. Our salience classification dataset is built using a novel method which grounds summary sentences to sentences in source with language model perplexity scores. Our approach (\figureref{fig:my_label}) resembles the \textit{extract-then-abstract} paradigm popular in prior work~\citep{gehrmann-etal-2018-bottom, liu2018generating, s2019extractive, Chen_2018}.\\

\vspace{-0.1in}


\noindent\textbf{Extraction Stage:}
To extract the most important content from the source document required to generate the summary,
we pose content selection as a binary classification task, labeling every sentence in the source document as \emph{salient} or \emph{non-salient}. Sentences classified as salient are concatenated in the order of occurrence in the source document\footnote{Maintaining the order of sentences ensures the logical flow of information is not disrupted.} to generate a compressed ``extractive summary'', which is then fed to the abstractive summarizer. In addition to identifying important information, the salience classifier is able to remove repetitive boilerplate text which is common in technical documents but often irrelevant to the actual content.\\
\vspace{-0.1in}

\noindent\textbf{Training Data for Salience Classification:} Since we do not have sentence-level training data for the classifier, we \emph{construct} it by grounding sentences of the ground truth summary to sentences in the source document. Consider a source document $S$ consisting of $m$ sentences $s_{1:m}$ and a target summary $T$ consisting of $n$ sentences  $t_{1:n}$ where $m>>n$. We compute the salience score for every source sentence $s_i \in S$ as $\frac{1}{n} \sum_{j = 0}^n f(s_i, t_j)$. Here $f(s, t)$ is a measure of how much source sentence $s$ grounds target sentence $t$. Following this, we sort the sentences in the source document based on salience score. The highest scoring 3$n$
sentences are chosen as \textit{salient} sentences and the lowest scoring 3$n$ are chosen as \textit{non-salient} sentences.\footnote{$3n$ is a tuned hyperparameter. Whenever $m < 6n$, we sort the sentences according to the salience score and assign \textit{salient} to the top half and \textit{non-salient} to the bottom half.} We construct our dataset for salience classification by running this algorithm for every $(S, T)$ pair in the training dataset. To ensure generalization with limited training data, we incorporate transfer learning and build our classifier by finetuning BERT-base~\citep{devlin-etal-2019-bert} using \texttt{transformers}~\citep{Wolf2019HuggingFacesTS}. More details on training are provided in \appendixref{appendix:classifier-details}.\\
\vspace{-0.1in}

\noindent\textbf{Choice of} $\boldsymbol{f(s, t)}$: To measure how much a source sentence $s$ grounds a target sentence $t$ we measure the perplexity of $t$ conditioned on $s$, using a pretrained language model GPT-2 large~\citep{radford2019language}. More formally, we concatenate $s$ and $t$ as $[s;~t]$ and feed it as input to GPT-2 large, calculating perplexity over the tokens of $t$. Here, \emph{a lower perplexity corresponds to a higher $f(s, t)$ score}. We find that this measure correlates with entailment and outperforms other choices of $f(s, t)$ like $n$-gram overlap, sentence embedding similarity \& entailment classifiers (\sectionref{sec:choice_of_lm}).\\ 

\vspace{-0.1in}


\noindent\textbf{Abstraction Stage:} Having compressed the source document using our extractor, we use a black-box pretrained abstractive summarizer trained on a related domain. In this work, we make use of the state-of-the-art model (i.e. BART), which is based on pretrained language models. Pretraining on CNN/DM helps BART generalize  to unseen but related domains like legal briefs.\footnote{Details on our BART setup are provided in \appendixref{appendix:bart_details}.}


\section{Experiments}
\label{sec:experiments}



\subsection{Evaluating the extractor}
\label{sec:extractor_eval}

To evaluate our proposed extractor, we first check whether our salience classifier generalizes to a held-out test set\footnote{Classifier data statistics at salient/non-salient sentences level: (Train=5363, Dev=1870, Test=2070)}. Indeed, it achieves a classification accuracy of 73.66\%, and qualitative analysis of the classifications confirm its ability to identify boilerplate sentences as \emph{non-salient}. Our classifier compresses source documents by 61\% on average.\footnote{Note that classifier score can be thresholded to obtain more or less compression depending on domain and end-task.} 
\\Next, we evaluate the quality of extracted salient sentences by checking the extent to which they overlap in information with the gold test set summaries, by measuring ROUGE-1/2 recall scores. As shown in \tableref{tab:recall}, our extractor outperforms a random selection of the same number of sentences and is comparable to the upper-bound recall performance achieved by feeding in the whole source document. Finally, to measure the extent to which our salience classifier matches human judgement, domain experts identified 8-10 salient sentences in four test documents with more than 200 sentences each on request. Despite their scarcity, our salience classifier recovers 64.7\% marked sentences, confirming correlation with human judgments.

\begin{table}[t]
\small
\centering
\begin{tabular}{lrr}
\toprule
Source & R-1 (Recall) & R-2 (Recall) \\
\midrule
Whole Document & 87.75        & 50.67        \\
Random Extractor & 78.66        & 38.53   \\
Proposed Extractor & 81.78        & 43.96        \\
\bottomrule
\end{tabular}
\caption{ROUGE-1/2 (R-1/2) recall scores of the gold summary with respect to the the ``Source'' document. Our saliency-driven extractor performs better than a random selection of the same number of sentences and is close to the upperbound recall performance achieved by feeding in the whole source document.}
\vspace{-0.1in}
\label{tab:recall}
\vspace{-0.1in}
\end{table}

\subsection{Evaluating the entire pipeline}
\label{sec:abstractive-eval}
We evaluate the entire pipeline by measuring the quality of abstractive summaries, obtained by feeding the extractive summary to BART. We study two abstractor settings: (1) Treating BART as a black-box with no modification; (2) Finetuning BART on the training and validation split of Amicus dataset\footnote{The training and validation splits together comprise of 96 documents. The test split was not used.}. We present results on the Amicus test set. We compare our model against several competitive baselines --- \textbf{(1) NE}: no extraction; \textbf{(2) Random}: a random selection of the same number of sentences as our extractive summary; \textbf{(3) TextRank}~\citep{mihalcea-tarau-2004-textrank, liu2018generating}: unsupervised graph based approach to rank text chunks within a document; \textbf{(4) Bottom-up summarizer}~\citep{gehrmann-etal-2018-bottom}: a strong \emph{extract-then-abstract} baseline where content selection is posed as a word-level sequence tagging problem. Similar to our setting, their content selector also uses large pretrained models (ELMo,~\citealp{peters-etal-2018-deep}), which we finetune on our training set. 




\begin{table}[t]
\centering
\small
\begin{tabular}{llrrr}
\toprule
Extractor &Abstractor                              & \multicolumn{1}{l}{R-1} & \multicolumn{1}{l}{R-2} & \multicolumn{1}{l}{R-L} \\
\midrule
NE &BART                                & 40.17                   & 13.36                   & 17.95                   \\
Random & BART            & 41.96                   & 13.30                   & 17.91                   \\
TextRank & BART                     & 42.63                   & 13.09                   & 17.93                   \\
Bottom-up & BART                    & 42.41                   & 14.50                   & 20.76                   \\
Ours & BART           & \textbf{44.97}                   & \textbf{15.37}                   & \textbf{23.95}                   \\
\midrule
NE &f.t. BART                      & 43.47                   & 16.30                   & 19.35                   \\
Random & f.t. BART            & 44.63              & 15.11                & 18.57                 \\
TextRank & f.t. BART                     & 45.10             & 15.51                   & 18.74               \\
Bottom-up & f.t. BART                    &44.89                  & 17.26                & 23.40                  \\
Ours & f.t. BART & \textbf{47.07}                   & \textbf{17.64}                   & \textbf{24.40}      \\
\bottomrule
\end{tabular}
\caption{Comparison of our method on the Amicus dataset with strong baselines. Our method outperforms all baselines in both Abstractor settings: (1) a pretrained CNN/DM BART; (2) the pretrained CNN/DM BART finetuned on the Amicus dataset (f.t. BART).} 
\vspace{-0.1in}
\label{table:amicus}
\end{table}

 As seen in \tableref{table:amicus}, we observe a 4.8 / 6 ROUGE-1/L improvement when compared to the no extractor baseline (NE), and 2.3 / 3.2 R-1/L improvement over the strongest extractor baseline (per metric); confirming the effectiveness of our method. In addition, finetuning the CNN/DM pretrained BART on 96 Amicus documents helps in domain adaption and boosts the ROUGE scores of both baselines and our method (f.t. BART). Specifically, we observe a 2.1 / 0.5 R-1/L boost in performance and outperform the best baseline (per metric) by 2.0 / 1.0 R-1/L points. Our model's improvements are statistically significant (p-value$<$ 0.06) except for when comparing our extractor + f.t BART with Bottom-up + f.t BART, the p-value is 0.16 due to the small test set. Refer \appendixref{appendix:bart_details} for qualitative analysis of our proposed model's generations.
\begin{figure}
    \centering
    \includegraphics[scale=0.5]{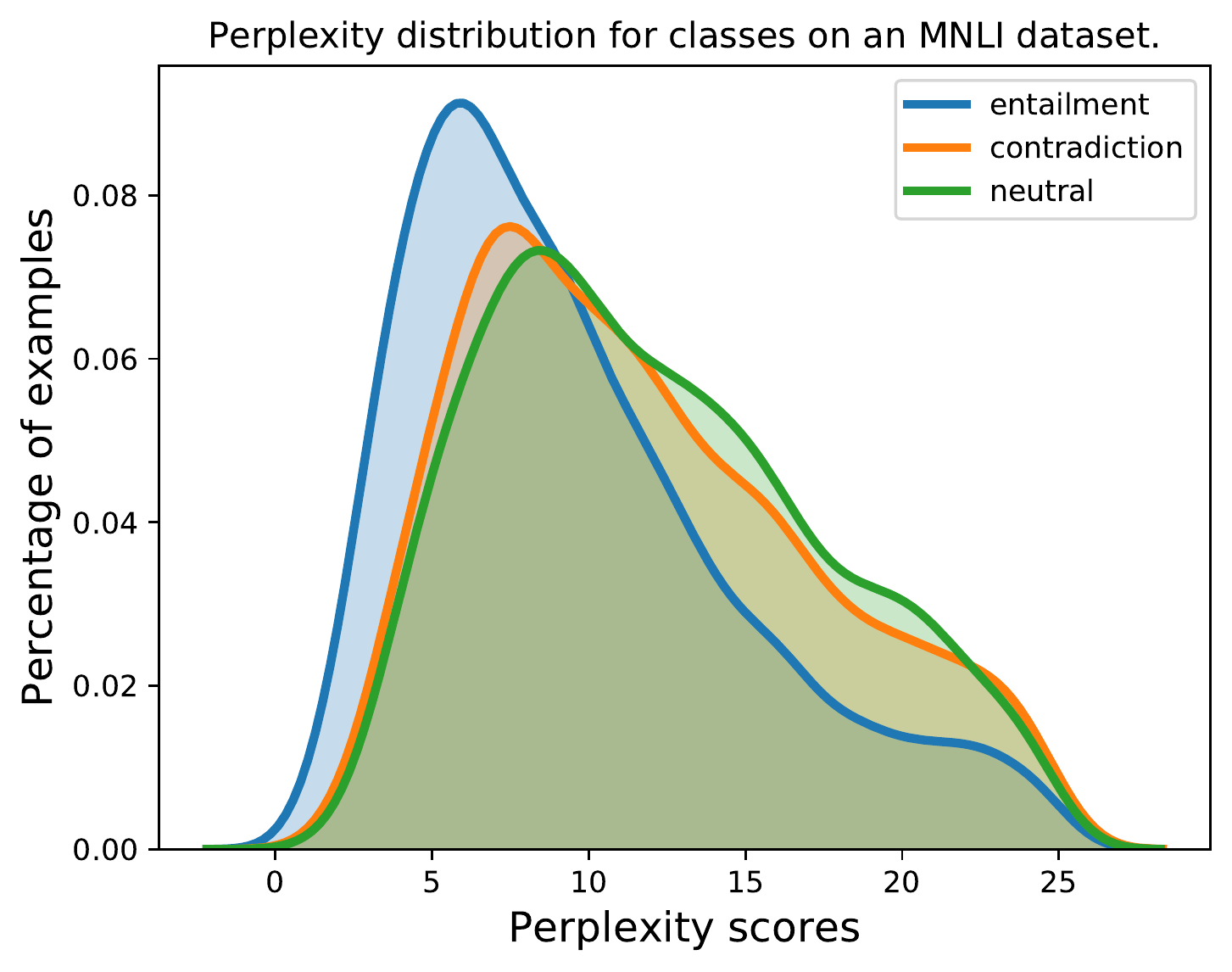}
    \vspace{-0.1in}
    \caption{Perplexity distribution of the hypothesis given the premise for each of the three classes sampled from the MultiNLI dataset. Entailment pairs tend to have lower perplexity, validating our choice of $f(s, t)$.}
    \vspace{-0.1in}
    \label{fig:perp}
\end{figure}

\begin{table}[t]
\small
\begin{tabular}{lrrr}
\toprule
Choice of $f(s,t)$          & R-1            & R-2            & R-L            \\
\midrule
Entailment (using RoBERTa) & 43.66          &  16.95          & 23.24          \\
Similarity (using BERT)    & 44.67          & 16.69          & 23.81           \\
BLEU (using nltk)          & 43.95          & 17.38          & 23.69          \\
Perplexity (using GPT-2)   & \textbf{47.07} & \textbf{17.64} & \textbf{24.40} \\
\bottomrule
\end{tabular}

\caption{Results of our \textit{extract-then-abstract} pipeline (after finetuning BART) by varying $f(s,t)$. Our choice of GPT-2 perplexity performs better than 3 alternatives.}
\vspace{-0.1in}
\label{tab:ablations}
\end{table}

\subsection{Validating the choice of $f(s, t)$}
\label{sec:choice_of_lm}

In \sectionref{sec:approach} we used GPT-2 perplexity scores to measure the extent to which a source sentence grounds a target sentence. To motivate this choice, we measure its correlation with existing entailment datasets. We randomly sample 5000 sentences from each class of the MultiNLI dataset~\citep{williams2018inference} and compute the perplexity of the hypothesis with the premise as context. As seen in \figureref{fig:perp}, entailment pairs tend to have the lowest perplexity. This motivates our choice of $f(s, t)$, since hypothesis sentences are best grounded in premise sentences for entailment pairs.\footnote{We hypothesize contradiction sentences have slightly lower perplexity than neutral due to more word overlap.}
To further validate the merit of GPT-2 perplexity, we conduct \textbf{ablations} using alternatives for $f(s, t)$: (1) Entailment score from a RoBERTa based MNLI classifier~\citep{liu2019roberta} (2) Cosine similarity of averaged embeddings from final layer of BERT~\citep{devlin-etal-2019-bert} (3) BLEU scores~\citep{papineni-etal-2002-bleu}. We present ROUGE scores using our whole \emph{extract-then-abstract} pipeline with different choices of $f(s,t)$ in \tableref{tab:ablations}. We note that perplexity performs the best, 2.4 ROUGE-1 
better than the best alternative and also performs 3.41 ROUGE-1 better than entailment.\footnote{We hypothesize that RoBERTa overfits on the MNLI dataset that also has known biases \citep{Gururangan_2018}.}



\section{Conclusion}
We tackle an important real-world problem of summarizing long domain-specific documents with very less training data. We propose an \textit{extract-then-abstract} pipeline which uses GPT-2 perplexity and a BERT classifier to estimate sentence salience. This sufficiently compresses a document, allowing us to use a pretrained model (BART) to generate coherent \& fluent summaries.
\bibliographystyle{acl_natbib}
\bibliography{eacl2021}

\begin{thebibliography}{25}
\expandafter\ifx\csname natexlab\endcsname\relax\def\natexlab#1{#1}\fi

\bibitem[{Chen and Bansal(2018)}]{Chen_2018}
Yen-Chun Chen and Mohit Bansal. 2018.
\newblock \href {https://doi.org/10.18653/v1/p18-1063} {Fast abstractive
  summarization with reinforce-selected sentence rewriting}.
\newblock \emph{Proceedings of the 56th Annual Meeting of the Association for
  Computational Linguistics (Volume 1: Long Papers)}.

\bibitem[{Cheng and Lapata(2016)}]{cheng-lapata-2016-neural}
Jianpeng Cheng and Mirella Lapata. 2016.
\newblock \href {https://doi.org/10.18653/v1/P16-1046} {Neural summarization by
  extracting sentences and words}.
\newblock In \emph{Proceedings of the 54th Annual Meeting of the Association
  for Computational Linguistics (Volume 1: Long Papers)}, pages 484--494,
  Berlin, Germany. Association for Computational Linguistics.

\bibitem[{Chopra et~al.(2016)Chopra, Auli, and
  Rush}]{chopra-etal-2016-abstractive}
Sumit Chopra, Michael Auli, and Alexander~M. Rush. 2016.
\newblock \href {https://doi.org/10.18653/v1/N16-1012} {Abstractive sentence
  summarization with attentive recurrent neural networks}.
\newblock In \emph{Proceedings of the 2016 Conference of the North {A}merican
  Chapter of the Association for Computational Linguistics: Human Language
  Technologies}, pages 93--98, San Diego, California. Association for
  Computational Linguistics.

\bibitem[{Cohan et~al.(2018)Cohan, Dernoncourt, Kim, Bui, Kim, Chang, and
  Goharian}]{cohan-etal-2018-discourse}
Arman Cohan, Franck Dernoncourt, Doo~Soon Kim, Trung Bui, Seokhwan Kim, Walter
  Chang, and Nazli Goharian. 2018.
\newblock \href {https://doi.org/10.18653/v1/N18-2097} {A discourse-aware
  attention model for abstractive summarization of long documents}.
\newblock In \emph{Proceedings of the 2018 Conference of the North {A}merican
  Chapter of the Association for Computational Linguistics: Human Language
  Technologies, Volume 2 (Short Papers)}, pages 615--621, New Orleans,
  Louisiana. Association for Computational Linguistics.

\bibitem[{Devlin et~al.(2019)Devlin, Chang, Lee, and
  Toutanova}]{devlin-etal-2019-bert}
Jacob Devlin, Ming-Wei Chang, Kenton Lee, and Kristina Toutanova. 2019.
\newblock \href {https://doi.org/10.18653/v1/N19-1423} {{BERT}: Pre-training of
  deep bidirectional transformers for language understanding}.
\newblock In \emph{Conference of the North American Chapter of the Association
  for Computational Linguistics}.

\bibitem[{Gehrmann et~al.(2018)Gehrmann, Deng, and
  Rush}]{gehrmann-etal-2018-bottom}
Sebastian Gehrmann, Yuntian Deng, and Alexander Rush. 2018.
\newblock \href {https://doi.org/10.18653/v1/D18-1443} {Bottom-up abstractive
  summarization}.
\newblock In \emph{Proceedings of Empirical Methods in Natural Language
  Processing}.

\bibitem[{Gu et~al.(2016)Gu, Lu, Li, and Li}]{gu-etal-2016-incorporating}
Jiatao Gu, Zhengdong Lu, Hang Li, and Victor~O.K. Li. 2016.
\newblock \href {https://doi.org/10.18653/v1/P16-1154} {Incorporating copying
  mechanism in sequence-to-sequence learning}.
\newblock In \emph{Proceedings of the 54th Annual Meeting of the Association
  for Computational Linguistics (Volume 1: Long Papers)}, pages 1631--1640,
  Berlin, Germany. Association for Computational Linguistics.

\bibitem[{Gururangan et~al.(2018)Gururangan, Swayamdipta, Levy, Schwartz,
  Bowman, and Smith}]{Gururangan_2018}
Suchin Gururangan, Swabha Swayamdipta, Omer Levy, Roy Schwartz, Samuel Bowman,
  and Noah~A. Smith. 2018.
\newblock \href {https://doi.org/10.18653/v1/n18-2017} {Annotation artifacts in
  natural language inference data}.
\newblock \emph{Proceedings of the 2018 Conference of the North American
  Chapter of the Association for Computational Linguistics: Human Language
  Technologies, Volume 2 (Short Papers)}.

\bibitem[{Lewis et~al.(2020)Lewis, Liu, Goyal, Ghazvininejad, Mohamed, Levy,
  Stoyanov, and Zettlemoyer}]{lewis2019bart}
Mike Lewis, Yinhan Liu, Naman Goyal, Marjan Ghazvininejad, Abdelrahman Mohamed,
  Omer Levy, Ves Stoyanov, and Luke Zettlemoyer. 2020.
\newblock Bart: Denoising sequence-to-sequence pre-training for natural
  language generation, translation, and comprehension.
\newblock In \emph{Proceedings of the Association for Computational
  Linguistics}.

\bibitem[{Liu et~al.(2018)Liu, Saleh, Pot, Goodrich, Sepassi, Kaiser, and
  Shazeer}]{liu2018generating}
Peter~J. Liu, Mohammad Saleh, Etienne Pot, Ben Goodrich, Ryan Sepassi, Lukasz
  Kaiser, and Noam Shazeer. 2018.
\newblock \href {http://arxiv.org/abs/1801.10198} {Generating wikipedia by
  summarizing long sequences}.

\bibitem[{Liu et~al.(2019)Liu, Ott, Goyal, Du, Joshi, Chen, Levy, Lewis,
  Zettlemoyer, and Stoyanov}]{liu2019roberta}
Yinhan Liu, Myle Ott, Naman Goyal, Jingfei Du, Mandar Joshi, Danqi Chen, Omer
  Levy, Mike Lewis, Luke Zettlemoyer, and Veselin Stoyanov. 2019.
\newblock Roberta: A robustly optimized bert pretraining approach.
\newblock \emph{arXiv preprint arXiv:1907.11692}.

\bibitem[{Magooda and Litman(2020)}]{magooda2020abstractive}
Ahmed Magooda and Diane Litman. 2020.
\newblock \href {http://arxiv.org/abs/2002.03407} {Abstractive summarization
  for low resource data using domain transfer and data synthesis}.

\bibitem[{Mihalcea and Tarau(2004)}]{mihalcea-tarau-2004-textrank}
Rada Mihalcea and Paul Tarau. 2004.
\newblock \href {https://www.aclweb.org/anthology/W04-3252} {{T}ext{R}ank:
  Bringing order into text}.
\newblock In \emph{Proceedings of the 2004 Conference on Empirical Methods in
  Natural Language Processing}, pages 404--411, Barcelona, Spain. Association
  for Computational Linguistics.

\bibitem[{Nallapati et~al.(2016)Nallapati, Zhou, dos Santos, Gulcehre, and
  Xiang}]{nallapati-etal-2016-abstractive}
Ramesh Nallapati, Bowen Zhou, Cicero dos Santos, Caglar Gulcehre, and Bing
  Xiang. 2016.
\newblock \href {https://doi.org/10.18653/v1/K16-1028} {Abstractive text
  summarization using sequence-to-sequence {RNN}s and beyond}.
\newblock In \emph{Proceedings of The 20th {SIGNLL} Conference on Computational
  Natural Language Learning}, pages 280--290, Berlin, Germany. Association for
  Computational Linguistics.

\bibitem[{Papineni et~al.(2002)Papineni, Roukos, Ward, and
  Zhu}]{papineni-etal-2002-bleu}
Kishore Papineni, Salim Roukos, Todd Ward, and Wei-Jing Zhu. 2002.
\newblock \href {https://doi.org/10.3115/1073083.1073135} {{B}leu: a method for
  automatic evaluation of machine translation}.
\newblock In \emph{Proceedings of the 40th Annual Meeting of the Association
  for Computational Linguistics}, pages 311--318, Philadelphia, Pennsylvania,
  USA. Association for Computational Linguistics.

\bibitem[{Parida and Motlicek(2019)}]{parida-motlicek-2019-abstract}
Shantipriya Parida and Petr Motlicek. 2019.
\newblock \href {https://doi.org/10.18653/v1/D19-1616} {Abstract text
  summarization: A low resource challenge}.
\newblock In \emph{Proceedings of the 2019 Conference on Empirical Methods in
  Natural Language Processing and the 9th International Joint Conference on
  Natural Language Processing (EMNLP-IJCNLP)}, pages 5994--5998, Hong Kong,
  China. Association for Computational Linguistics.

\bibitem[{Peters et~al.(2018)Peters, Neumann, Iyyer, Gardner, Clark, Lee, and
  Zettlemoyer}]{peters-etal-2018-deep}
Matthew Peters, Mark Neumann, Mohit Iyyer, Matt Gardner, Christopher Clark,
  Kenton Lee, and Luke Zettlemoyer. 2018.
\newblock \href {https://doi.org/10.18653/v1/N18-1202} {Deep contextualized
  word representations}.
\newblock In \emph{Proceedings of the 2018 Conference of the North {A}merican
  Chapter of the Association for Computational Linguistics: Human Language
  Technologies, Volume 1 (Long Papers)}.

\bibitem[{Radford et~al.(2019)Radford, Wu, Child, Luan, Amodei, and
  Sutskever}]{radford2019language}
Alec Radford, Jeffrey Wu, Rewon Child, David Luan, Dario Amodei, and Ilya
  Sutskever. 2019.
\newblock Language models are unsupervised multitask learners.
\newblock \emph{OpenAI Blog}, 1(8):9.

\bibitem[{Rush et~al.(2015)Rush, Chopra, and Weston}]{Rush_2015}
Alexander~M. Rush, Sumit Chopra, and Jason Weston. 2015.
\newblock \href {https://doi.org/10.18653/v1/d15-1044} {A neural attention
  model for abstractive sentence summarization}.
\newblock \emph{Proceedings of the 2015 Conference on Empirical Methods in
  Natural Language Processing}.

\bibitem[{See et~al.(2017)See, Liu, and Manning}]{See_2017}
Abigail See, Peter~J. Liu, and Christopher~D. Manning. 2017.
\newblock \href {https://doi.org/10.18653/v1/p17-1099} {Get to the point:
  Summarization with pointer-generator networks}.
\newblock \emph{Proceedings of the 55th Annual Meeting of the Association for
  Computational Linguistics (Volume 1: Long Papers)}.

\bibitem[{Subramanian et~al.(2019)Subramanian, Li, Pilault, and
  Pal}]{s2019extractive}
Sandeep Subramanian, Raymond Li, Jonathan Pilault, and Christopher Pal. 2019.
\newblock \href {http://arxiv.org/abs/1909.03186} {On extractive and
  abstractive neural document summarization with transformer language models}.

\bibitem[{Williams et~al.(2018)Williams, Nangia, and
  Bowman}]{williams2018inference}
Adina Williams, Nikita Nangia, and Samuel Bowman. 2018.
\newblock \href {http://aclweb.org/anthology/N18-1101} {A broad-coverage
  challenge corpus for sentence understanding through inference}.
\newblock In \emph{Conference of the North American Chapter of the Association
  for Computational Linguistics}.

\bibitem[{Wolf et~al.(2019)Wolf, Debut, Sanh, Chaumond, Delangue, Moi, Cistac,
  Rault, Louf, Funtowicz, and Brew}]{Wolf2019HuggingFacesTS}
Thomas Wolf, Lysandre Debut, Victor Sanh, Julien Chaumond, Clement Delangue,
  Anthony Moi, Pierric Cistac, Tim Rault, R'emi Louf, Morgan Funtowicz, and
  Jamie Brew. 2019.
\newblock Huggingface's transformers: State-of-the-art natural language
  processing.
\newblock \emph{ArXiv}, abs/1910.03771.

\bibitem[{Wu and Hu(2018)}]{wu2018learning}
Yuxiang Wu and Baotian Hu. 2018.
\newblock \href {http://arxiv.org/abs/1804.07036} {Learning to extract coherent
  summary via deep reinforcement learning}.

\bibitem[{Yogatama et~al.(2019)Yogatama, d'Autume, Connor, Kocisky,
  Chrzanowski, Kong, Lazaridou, Ling, Yu, Dyer et~al.}]{yogatama2019learning}
Dani Yogatama, Cyprien de~Masson d'Autume, Jerome Connor, Tomas Kocisky, Mike
  Chrzanowski, Lingpeng Kong, Angeliki Lazaridou, Wang Ling, Lei Yu, Chris
  Dyer, et~al. 2019.
\newblock Learning and evaluating general linguistic intelligence.
\newblock \emph{arXiv preprint arXiv:1901.11373}.

\end{thebibliography}
\newpage
~
\appendix

\section{Appendix}
\subsection{Data pre-processing}
In this section, the various pre-processing steps of data performed at different stages are explained.\\

\textbf{Extracting (document, summary) pairs}: The 120 pairs of Amicus Briefs were scrapped from their website\footnote{ https://publichealthlawcenter.org/amicus-briefs}. The Summary of Arguments section of the Amicus Briefs was extracted as the target summary and the main content excluding title page, table of contents, acknowledgements, appendix etc was extracted as document/source.\\

\textbf{Sentence pre-processing}: Sentences from the (document, summary) files were split using the spaCy\footnote{https://pypi.org/project/spacy/} sentence splitter. Furthermore, the sentences were each processed to remove special characters using regex rules. If a sentence contained less that 5 words, it was pruned out from the computation of $f(s,t)$ to reduce the complexity of pairs considered.  

\subsection{Sentence Saliency Classifier}
\label{appendix:classifier-details}

\textbf{Training Details}: Our classifier uses the BERT sequence labeling configuration\footnote{\url{https://huggingface.co/transformers/model_doc/bert.html##transformers.BertForSequenceClassification}} from \texttt{transformers}~\citep{Wolf2019HuggingFacesTS}, which is a pretrained BERT-base model with an initially untrained classification head on the \texttt{[CLS]} feature vector. This model is then finetuned for 5 epochs using the training data which consists of 5363 sentences in the Amicus dataset (equal distribution among the two classes). We use a train / dev / test split of 60\%, 20\%, 20\%.  Training configuration of the classifier is as follows: learning rate = 2e-5, max\_grad\_norm = 1.0, num\_training\_steps = 1000, num\_warmup\_steps = 100, warmup\_proportion = 0.1, Optimizer = Adam, Schduler = linear with warmup.\\

\textbf{Alternate methods to choose +/- samples}: The aggregate scoring method mentioned in \sectionref{sec:approach} was one choice to pick salient and non-salient samples for each document. Aggregate method compresses the source by 61\% on an average. The other methods experimented were:
        \begin{itemize}
            \item Top k - Bottom k: $\forall t_j \in \text{T}$, we picked the top-k scoring source sentences as positive samples and the bottom-k sentences as the negative samples ensuring that \{positive\} $\cap$ \{negative\} $=0$. Using this technique, the classifier achieves accuracy of nearly 1 as can be seen from \tableref{tab:classifier-all-combinations}. On qualitative analysis,  we identified that there is a clear distinction in the positive and the negative examples. Eg: sentences such as `This document is prepared by XYZ' would be picked as non salient sentence and classifier is able to achieve high accuracy. This could however be used to train a classifier to \emph{identify boiler plate sentences} across the document. This method compresses source document by 63\% on an average.
            \item Random negative sampling: Salient examples were chosen for a document as per the above method. For the non salient examples, we randomly sampled from the rest of the document. This allows the classifier to learn about sentences that that are difficult to be classified as positive or negative. Hence, the accuracy of the classifier is lower than the other two methods as can be seen from \tableref{tab:classifier-all-combinations}. This method compresses the source document by 70\% on an average.
        \end{itemize}
        
\textbf{Compute time and resources}: Execution time for different choice of \emph{f(s,t)} for all 120 pairs:
\begin{itemize}
    \item Perplexity using GPT-2:executes within 15hrs using 2 GPUs
    \item Entailment score using RoBERTa: executes within 22hrs using 2 GPUs
    \item Cosine Similarity using BERT [CLS] embeddings: executes within 3hrs on a single GPU
    \item BLEU score using nltk: executes within 15min on a single GPU.
    \end{itemize}
These scores need to be generated once and can be reused for various experiments. Sampling methods to choose salient and non-salient sentences for each document takes less than a minute to run. \\

\begin{table}[ht!]
\small
\centering
\begin{tabular}{lll}
\toprule
Sampling Method                                                                                                                                & f(s,t) & Accuracy \\ \toprule
\multirow{4}{*}{\begin{tabular}[c]{@{}l@{}}Aggregate scoring\\ for each source\\ sentence.\end{tabular}}                                                   & BLEU            & 0.7813            \\
                                                                                                                                                           & Perplexity      & 0.7366            \\
                                                                                                                                                           & Entailment      & 0.6569            \\
                                                                                                                                                           & Similarity      & 0.8391            \\ \hline
\multirow{4}{*}{\begin{tabular}[c]{@{}l@{}}Top k-Bottom k \\ sources sentences\\ or each summary sentence\end{tabular}}                                    & BLEU            & 0.9997            \\
                                                                                                                                                           & Perplexity      & 0.9915            \\
                                                                                                                                                           & Entailment      & 0.9973            \\
                                                                                                                                                           & Similarity      & 1                 \\ \hline
\multirow{4}{*}{\begin{tabular}[c]{@{}l@{}}Top k for each summary \\ sentence and random \\ negative sampling from \\ the remaining document.\end{tabular}} & BLEU            & 0.5784            \\
                                                                                                                                                           & Perplexity      & 0.655             \\
                                                                                                                                                           & Entailment      & 0.5611            \\
                                                                                                                                                           & Similarity      & 0.6233            \\ \bottomrule
\end{tabular}
\caption{The accuracy of the held out set of Amicus for different classifiers trained on the data prepared using choice of different f(s,t) and sampling methods.  Here, k=3. }
\label{tab:classifier-all-combinations}
\end{table}

\textbf{Analysis}: 
(a) \tableref{tab:classifier-all-combinations} shows the classifier accuracies for combinations of \emph{f(s,t)} and sampling methods. We observe that for the aggregate sampling method, although perplexity based classifier does not have the highest accuracy, our pipeline where $f(s,t)$ is perplexity score gives the best result(ROUGE) amongst the ablation experiments(\tableref{tab:ablations}). Classifier accuracy is determined on automated labelling based on the saliency score, rather than true labels, hence best classifier does not imply best summarization. (b) \tableref{tab:perplexity-examples} shows the examples of using perplexity as \emph{f(s,t)} to see how the summary grounds the source. The table shows three summary sentences and the corresponding source sentences that had the lowest perplexity scores. We can see that, summary either has a similar meaning or logically follows the source. (c) \tableref{tab:posneg-examples} has three examples each for salient sentences and non-salient sentences inferred by the classifier trained on data prepared as mentioned in \sectionref{sec:approach}. The third sentence in the non-salient sentences column is an example of boiler-plate content detected that is present across documents. 

\subsection{Abstractive Summarizer: BART}
\label{appendix:bart_details}

BART is a seq2seq model based on denoising pretraining objective which is supposed to generalize better on various natural language understanding tasks; abstractive summarization being one of them. For abstractive stage of our proposed approach, we decided to see (\emph{bart.large.cnn}) variant which is essentially BART-large model (with 12 encoder and decoder layers and 400 million parameters) finetuned for CNN/DM summarization task. We use the pre-computed weights available for use here\footnote{\url{https://github.com/pytorch/fairseq/tree/master/examples/bart}}. Using BART's text generation script, we set length penalty (lenpen) as 2.0 and minimum length (min\_len) as 500 words in order to encourage BART to produce longer outputs which is more suitable to our dataset. Also, we use beam size of 4 and and no\_repeat\_ngram\_size of 3.\\

\noindent\textbf{Finetuning:} We use the train and dev splits of Amicus dataset (96 source-target pairs) and finetune BART for summarization task starting from its CNN/DM finetuned checkpoint. First, we pre-process the dataset as per the guidelines in the official code\footnote{\url{https://github.com/pytorch/fairseq/blob/master/examples/bart/README.summarization.md}}. We finetune for 500 epochs with learning rate of 3e-5 and early stop if validation loss doesn't decrease for 50 epochs. Others parameters are as follows: total\_num\_updates = 20000, warmup\_updates = 500, update\_freq = 4, optimiser = Adam with weight decay of 0.01. Rest of parameters were kept as default in the official script. Results (Precision, Recall, F1) on the test set of Amicus using the existing BART model and finetuned BART are shown in \tableref{tab:full}.  \\

\tableref{tab:summaries} shows an example of target summary and  summary generated by our model(\sectionref{sec:approach}) for one sample source document. We can see that the summary generated by our model is fluent and has coherent flow of information.

\begin{table*}[t]
\hspace*{-0.2in}
\centering
\small
\begin{tabular}{|l|l|}
\hline
Summary Sentence                                                                                                                                                                                   & Source Sentence                                                                                                                                                                                                                                                                                                                                                                                                \\ \hline
\begin{tabular}[c]{@{}l@{}}In the immigration context, this jurisprudence has\\ prompted the Court to reject the notion that  \\ the so-called entry fiction is of constitutional significance.\end{tabular} & \begin{tabular}[c]{@{}l@{}}Prior to Knauff and Mezei, the distinction \\ between noncitizens who had entered the\\ United States and those who remained outside \\ it had not had been elevated to a bright-line constitutional\\ rule, and entry had never been completely determinative\\ of the fact or extent of protection under the Due \\ Process Clause.\end{tabular}                                            \\ \hline
\begin{tabular}[c]{@{}l@{}}It has accordingly authorized such detention only in limited\\ circumstances pursuant to a carefully defined scheme.\end{tabular}                                                 & \begin{tabular}[c]{@{}l@{}}The Court's substantive due process jurisprudence also\\ recognizes that an individual may be subjected to regulatory\\ detention only in narrow circumstances under a carefully\\ drawn scheme.\end{tabular}                                                                                                                                                                                 \\ \hline
\begin{tabular}[c]{@{}l@{}}With respect to substantive due process, this Court has\\ increasingly recognized the punitive consequences of indefinite \\ regulatory detention.\end{tabular}                   & \begin{tabular}[c]{@{}l@{}}Thus, the Court has substantially restricted the availability and\\ duration of regulatory confinement in the | years since it decided \\ Meze1.In Zadvydas, this Court established that its substantive \\ due process jurisprudence provided the appropriate framework\\ for evaluating the administrative detention of noncitizens \\ pending removal from the United States.\end{tabular} \\ \hline
\end{tabular}
\caption{Using GPT-2 perplexity as f(s,t), here are three sentences from the summary with corresponding source sentence, having the lowest perplexity. }
\label{tab:perplexity-examples}
\end{table*}
\begin{table*}[ht!]
\small
\begin{tabular}{|l|l|}
\hline
Salient Sentences                                                                                                                                                                                                                                                                                                & Non-Salient sentences                                                                                                                                                                              \\ \hline
\begin{tabular}[c]{@{}l@{}}The same time, the Court has long been skeptical of the \\ military's authority to try individuals other than \\ active service personnel.\end{tabular}                                                                                                                                         & \begin{tabular}[c]{@{}l@{}}A government predicated on checks and balances serves\\ not only to make Government accountable but also to\\ secure individual liberty.\end{tabular}                            \\ \hline
\begin{tabular}[c]{@{}l@{}}On the basis of this revised test, the Court of Appeals\\ refused to apply the exceptional circumstances exception\\ to Al-Nashiri's petition.\end{tabular}                                                                                                                                     & \begin{tabular}[c]{@{}l@{}}At present, the Rules for Courts-Martial require that the\\ accused be brought to trial within 120 days after\\ the earlier of preferral of charges or confinement.\end{tabular} \\ \hline
\begin{tabular}[c]{@{}l@{}}Consonant with that tradition,\\ this Court should review the Court of Appeals' \\ decision to confirm that exceptional delay before trial remains\\ of central concern on habeas review and is indeed one of the\\ very dangers the writ of habeas corpus was designed to avoid.\end{tabular} & \begin{tabular}[c]{@{}l@{}}Respectfully submitted, May 31, 2017 LINDA A. KLEIN \\ Counsel of Record AMERICAN BAR ASSOCIATION\\ 321 North Clark Street Chicago ...\end{tabular}                              \\ \hline
\end{tabular}
\caption{This table shows the sentences classified as salient and non-salient from one Amicus source document. We can see that the last sentence in the non-salient sentences column shows an example of boiler-plate content present across documents. The classifier is trained on data chosen on aggregate score of source sentences where \emph{f(s,t)} is GPT-2 perplexity.}
\label{tab:posneg-examples}
\end{table*}
\label{sec:appendix}

\begin{table*}[ht]
\centering
\small
\begin{tabular}{llrrrr}
\toprule
\multicolumn{2}{c}{Metric}           & \multicolumn{1}{l}{BART} & \multicolumn{1}{l}{Ours + BART} & \multicolumn{1}{l}{f.t. BART} & \multicolumn{1}{l}{Ours + f.t. BART} \\
\toprule
\multirow{3}{*}{ROUGE-1} & Recall    & 40.87                    & 47.46                           & 46.90                         & 56.04                                \\
                         & Precision & 47.21                    & 49.97                           & 48.68                         & 46.16                                \\
                         & F-1       & 40.17                    & 44.97                           & 43.47                         & 47.07                                \\
                         \hline
\multirow{3}{*}{ROUGE-2} & Recall    & 13.76                    & 16.54                           & 17.84                         & 21.50                                \\
                         & Precision & 15.46                    & 17.04                           & 17.84                         & 17.10                                \\
                         & F-1       & 13.36                    & 15.37                           & 16.30                         & 17.64                                \\
                         \hline
\multirow{3}{*}{ROUGE-L} & Recall    & 18.34                    & 25.58                           & 21.30                         & 29.62                                \\
                         & Precision & 21.04                    & 26.27                           & 21.35                         & 23.47                                \\
                         & F-1       & 17.95                    & 23.95                           & 19.35                         & 24.40    \\
                         \bottomrule
\end{tabular}
\caption{Overall pipeline results by adding our extractor (\emph{f(s,t)} as GPT-2 perplexity + Classifier) to BART and finetuned BART (f.t. BART), including the precision and recall values for each metric.}
\label{tab:full}
\end{table*}





\begin{table*}[t]
\centering
\small
\begin{tabular}{|c|}
\hline
\begin{tabular}[c]{@{}c@{}}This Court's determination of whether due process under the New HampshireConstitution requires\\ court-appointed counsel for indigent parent-defendants, in order to protect their fundamental right\\ to parent, requires the balancing of three factors--(1) theprivate interest at stake, (2) the risk of error\\ and the value of procedural safeguards, and (3)the state's interest. See In re Shelby R., 148 N.H. 237,\\ 240 (2002) (citing In re Richard A., 146 N.H..295, 298 (2001)). Because there is no dispute that the \\ fundamental right to parent isat stake in abuse and neglect proceedings, the ABA focuses its \\ discussion on the second and third factors of the three factor test.As to the second, so-called "risk of error"\\ factor, the ABA's conclusion, after years of investigation and analysis, is that the absence of counsel for \\ indigent parent-defendants in abuse and neglect proceedings results in a significant risk of an erroneous\\ determination. This is especially true where the opposing party is the State. As to the third, state's interest\\ factor, the ABA's investigation shows that the interests of both the parent and the state are best served\\ where indigent parent-defendants are represented. The ABA respectfully suggests that the evidence and\\ analysis relevant to these two factors is so compelling in most, if not all, abuse and neglect proceedings\\ involving indigent parent-defendants, that a case-by-case balancing of the factors should be rejected in \\ favor of a rule requiring the appointment of counsel] for indigent parent-defendants in all such proceedings.\\ The evidence and analysis supporting the ABA's policy includes the fact that a substantial majority of states \\ have recognized an unqualified right to counsel for indigentparent-defendants in child custody proceedings. \\ Similarly, other industrial democraciesprovide indigent parent-defendants with such right to counsel. The \\ ABA respectfully submits that this Court should require no less as a matter of due process under the New \\ Hampshire Constitution.Although of whetherJn re Shelby R. resulted in a or not a natural parent'splurality\\ role inruling, the the familyCourt is awas not split fundamentalon the libertyquestion interestprotected by \\ the State Constitution. See In re Shelby R., 148 NH. at 244 (dissenting opinion).\end{tabular}                                                                                                                                                                                                                                                                                                                                                                                                                      \\ \hline
\begin{tabular}[c]{@{}c@{}}Hampshire constitution requires this court to determine whether indigentparents have a legally protected interest.\\ Most indigent parent - defendants are incapable of performingthe advocacy functions required in abuse and \\ neglectproceedings. Most unrepresented parents cannot perform the advocacy functions - - including investigating\\ facts , making an orderly factual presentation , and cross - examining witnesses - - that are required. The intense,\\ emotionally charged backdrop against which custody decisionsare often made further exacerbates the inherent\\ disadvantages faced by unrepresented indigent parents. The need for counsel for the indigentparent - defendant is\\ especially great where the opposing party is the state. The court must weighthree factors : ( 1) the private interests \\ that will be affected. ( 2) the risk of erroneousdeprivation of the liberty interest through the procedures used and the\\ value , if any, ofadditional or substitute procedural safeguards. ( 3) the state ' s interest , including the function \\ involved and fiscal and administrative burdens that additional or substituteprocedural requirements would entail id\\ at 240 ; see also in re father , 155 n . h . 93 , 95 ( 2007 ) . this court has previously concluded as to the first factor\\ that adversary child custody proceedings implicate a fundamental liberty interest - - the right to parent in this case,\\ the central question thus becomes whether that right is sufficiently protected. The conclusion that counsel must be \\ provided is so compelling in most , if not all cases , that a case - by - case balancing of the factors should be rejected\\ in favor of a rule requiring the appointment of counsel for lowincome parent - defendant in all such proceedings to be \\ constitutionally acceptable. The state is not the only adversary finding the only meaningful right to be heard when her\\ adversary is not represented by counsel is not spaled against the traditional weapons of the state, such as the state’s\\ attorney general. The courts must also weigh the public interest in the child custody case, including the function \\ involved and the cost of additional or substitute safeguards, as well as the cost to the state of the additional or substituted\\ safeguards. The risk of an erroneous deprivation of the findamentalright to parent only increases the only increase in\\ the risk that the state will find the child is not heard when the state is the adversary. The public interest is only \\ increased by the fact that the child will not be heard by the state when the parent is represented by a lawyer. \\ The high level of complexity of child custody cases makes it difficult for the court to make a fair and just decision.\end{tabular} \\ \hline
\end{tabular}

\caption{The table shows the comparison of summaries where the top summary is the target summary and the bottom summary is the one generated by our extractor and f.t BART. As we can see, the summary is coherent and has fluent information flow. }
\label{tab:summaries}
\end{table*}

\end{document}